%% file: arxiv.tex
\documentclass[10pt,twocolumn,letterpaper]{article}

\usepackage[pagenumbers]{arxiv} 
\usepackage{booktabs}       
\usepackage{amsfonts}       
\usepackage{nicefrac}       
\usepackage{microtype}      
\usepackage{xcolor}         
\usepackage{multicol}
\usepackage{multirow}
\usepackage{booktabs}
\usepackage{caption}
\usepackage{subcaption}
\usepackage{tabularx}
\usepackage{newfloat}
\usepackage{listings}
\usepackage{tcolorbox}
\usepackage{algorithm}
\usepackage{algorithmic}
\usepackage{colortbl}
\usepackage{lscape}
\usepackage{graphicx}
\usepackage{pifont}
\usepackage{subcaption} 
\usepackage{wrapfig}
\usepackage{floatflt}
\usepackage{tcolorbox}
\usepackage{marvosym}
\usepackage{afterpage}
\usepackage{ulem}

\usepackage{mathtools}
\definecolor{cvprblue}{rgb}{0.21,0.49,0.74}
\definecolor{c1}{HTML}{F2C335}
\definecolor{c2}{HTML}{D9666F}
\definecolor{c3}{HTML}{6D339C}
\definecolor{c4}{HTML}{285D9B}
\definecolor{mygray}{HTML}{E6F0E8}
\definecolor{color1}{HTML}{ECF4F9}
\definecolor{color2}{HTML}{FFF1E0}
\definecolor{color3}{RGB}{236,244,249} 
\definecolor{Gray}{gray}{0.95}
\definecolor{LightBlue}{HTML}{FBEFF0}

\definecolor{vis1}{HTML}{C00000}
\definecolor{vis2}{HTML}{002060}
\definecolor{vis3}{HTML}{196B24}

\newcommand{\CC}[1]{\cellcolor{LightBlue}}
\newcommand{\RC}[1]{\rowcolor{LightBlue}}
\definecolor{mydeepgreen}{RGB}{0,100,0} 
\definecolor{deepred}{RGB}{139,0,0} 

\usepackage[pagebackref,breaklinks,colorlinks,allcolors=cvprblue]{hyperref}

\def\paperID{2853} 
\def\confName{CVPR}
\def\confYear{2026}

\hypersetup{
    colorlinks=true,
    urlcolor=c2,
}

\title{\textcolor{c2}{V}ideo-as-\textcolor{c4}{Ans}wer: Predict and Generate Next Video Event with Joint-GRPO}

\author{Junhao Cheng$^{1\dagger}$\qquad Liang Hou$^{2}$ \qquad Xin Tao$^{2}$ \qquad  Jing Liao$^{1}$\vspace{3pt}\\
\normalsize$^1$ City University of Hong Kong \qquad \normalsize$^2$ Kling Team, Kuaishou Technology \vspace{3pt}\\
\url{https://video-as-answer.github.io/}
\vspace{10pt}
}

\begin{document}
\maketitle
\renewcommand*{\thefootnote}{}
\footnotetext[1]{$^\dagger$This work was conducted during the author's internship at Kling Team, Kuaishou Technology.}

\input{sec/0_abstract}    
\input{sec/1_intro}

\input{sec/2_related_works}
\input{sec/3_dataset}
\input{sec/4_methods}

\input{sec/5_exp}
\input{sec/6_Conclusion}
\clearpage
\appendix
\input{sec/X_suppl}

{
    \small
    \bibliographystyle{ieeenat_fullname}
    \bibliography{main}
}


\end{document}

%% file: sec/0_abstract.tex
\begin{abstract}
While language models have become impactful in many real-world applications, video generation remains largely confined to entertainment. Motivated by video's inherent capacity to demonstrate physical-world information that is difficult to convey through language alone (e.g., imagine teaching someone to tie a tie using only text), we identify an underutilized opportunity to extend video as a new answer modality for Next-Event Prediction (NEP), formalized as \textbf{Video-Next-Event Prediction (VNEP)}. While the established NEP task takes a video with a procedural or predictive question as input to predict the next event in text, VNEP requires dynamic video responses. This shift from telling to showing unlocks more intuitive and customized answers for procedural learning and creative exploration. However, this task remains challenging for existing models, as it demands an understanding of multimodal input, instruction-conditioned reasoning, and the generation of video with visual and semantic consistency. To address this, we introduce \textbf{VANS}, a model that leverages reinforcement learning to align a Vision-Language Model (VLM) with a Video Diffusion Model (VDM) for VNEP. The core of VANS is our proposed \textbf{Joint-GRPO} that orchestrates the VLM and VDM to function as a unit. Driven by a shared reward on their respective output, it optimizes the VLM to produce captions that are both accurate and friendly to visualize, while guiding the VDM to generate videos that are faithful to these captions and the input visual context. To enable this learning, we craft \textbf{VANS-Data-100K}, a dedicated dataset for the VNEP task. Experiments on procedural and predictive benchmarks demonstrate that VANS achieves state-of-the-art performance in both video event prediction and visualization. Codes are released in \url{https://github.com/KlingTeam/VANS}.
\end{abstract}

%% file: sec/1_intro.tex
\section{Introduction}
\label{sec:intro}

While generative AI~\cite{yang2025qwen3,team2024gemini,liu2024deepseek,yang2024cogvideox,wan2025wan} has revolutionized text-based tasks in real-life domains like healthcare~\cite{goyal2024healai} and education~\cite{chu2025llm}, video generation models remain largely confined to entertainment~\cite{cheng2025animegamer,yu2025gamefactory}. This is a missed opportunity, as video encapsulates rich, dynamic information about the physical world that text alone struggles to convey~\cite{searle1980minds,yang2024video}.

\begin{figure}[!t]
	\centering
	\includegraphics[width=0.47\textwidth]{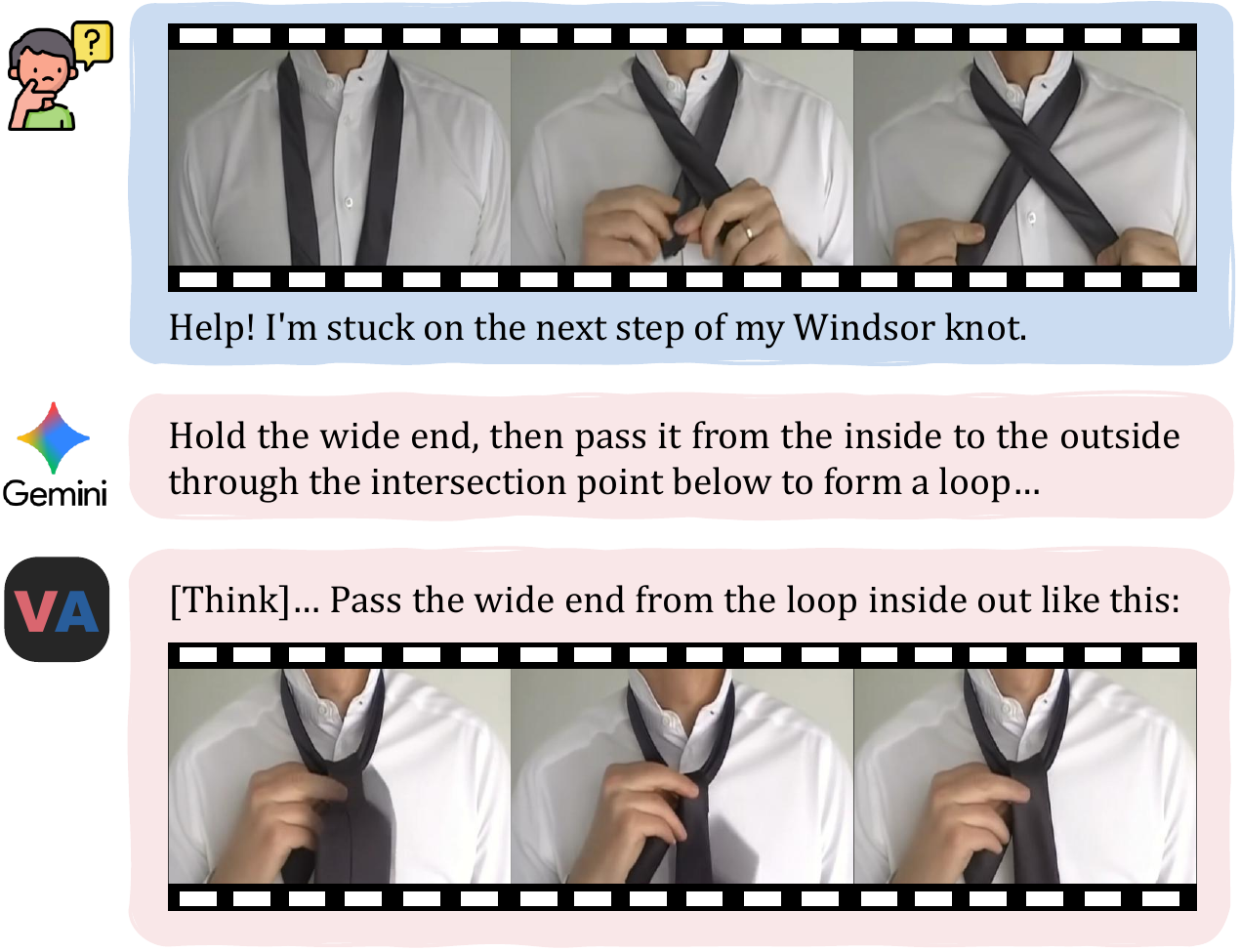}
\caption{Video answer (our VANS) versus text-only answer (Gemini) on a procedural question. Video answer provides an intuitive and customized response by demonstrating the action directly, while text-only answer falls short in clarity.}
\label{fig: intro taser}
\end{figure}

Motivated by this observation, we pioneer \textbf{Video-Next-Event Prediction (VNEP)}, a novel paradigm that uses video as the answer modality for Next-Event Prediction (NEP). While the established NEP task takes a video with a procedural or predictive question as input to predict the next event in text~\cite{lai2025cross,cheng2025TEMPURA,souvcek2025showhowto,wu2025stitch}, VNEP instead generates dynamic video responses. The shift from telling to showing enables VNEP to offer more intuitive and customized answers by leveraging video’s ability to convey spatial layout, motion, and temporal ordering, while adapting the demonstration to the user’s current state. As shown in Figure~\ref{fig: intro taser}, the video answer guides the user through the remaining steps of the Windsor knot based on his current tie configuration (e.g., color, orientation and tightness), providing clarity and personalization that a generic text-only description cannot achieve.

However, VNEP introduces significant challenges that go beyond mere visual continuation. Unlike existing video extension task~\cite{zhuang2025videogptclipdiffusion} which predicts frames based on spatiotemporal patterns (e.g., forecasting a ball's trajectory), VNEP focus on event-conditioned reasoning. It requires a model to first comprehend the video and question, reason about the subsequent event from causal or procedural logic (e.g., inferring that adding soap is needed after observing dirty dishes being scrubbed with water), and then generate a video that is both visually coherent and semantically faithful to this inferred event. A straightforward solution is to employ a VLM for prediction followed by a VDM for generation. However, this cascaded pipeline suffers from a semantic-to-visual misalignment; the VLM's textual output may be linguistically correct but visually unrealistic or unexecutable by the VDM, leading to semantically and visually divergent videos~\cite{xiao2025mindomni}. In contrast, unified models~\cite{luo2025univid,wei2025univideo,tan2025omni} attempt to align understanding and generation within a single model but face a capability trade-off, often excelling in one at the expense of the other and struggling to achieve optimal performance in both simultaneously~\cite{xiao2025haploomni}. Consequently, neither paradigm alone offers a satisfactory solution.

\begin{figure*}[!t]
	\centering
	\includegraphics[width=\textwidth]{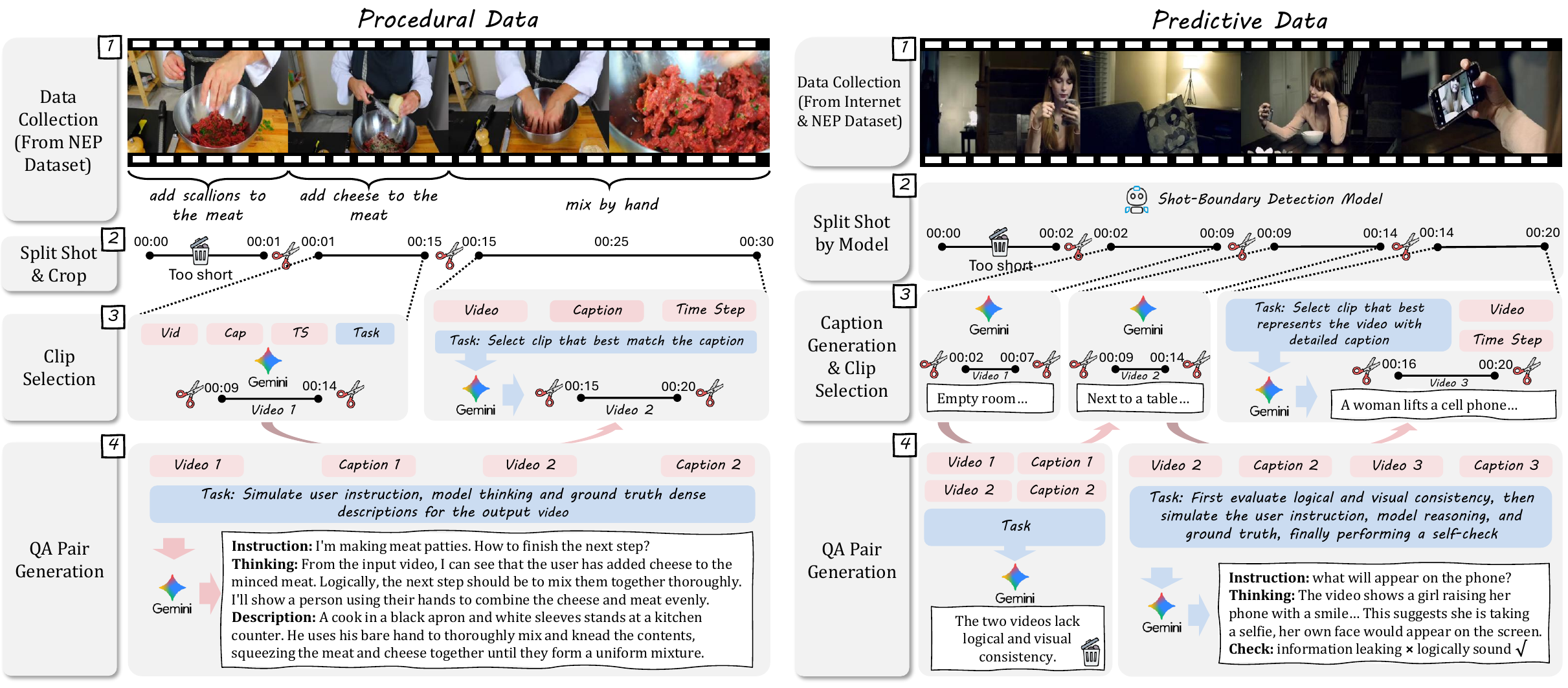}
\caption{Data curation pipeline of VANS-Data-100K, which processes raw videos through shot splitting, clip selection, and QA generation to produce high-quality data for both procedural and predictive Video-Next-Event Prediction.}
\label{fig: data pipeline}
\end{figure*}

However, the limitations of these two paradigms point toward a more promising direction: a tight integration of specialized models that preserves their strengths while resolving their interoperability issues. To this end, we propose \textbf{VANS}, a model that employs reinforcement learning (RL) post-training as an effective alignment process to fully realize the complementary strengths of VLMs (in semantic reasoning) and VDMs (in visual synthesis), enabling them to operate in concert for VNEP. Central to our approach is the \textbf{Joint-GRPO} RL strategy, which orchestrates both models using a joint reward derived from their respective outputs. Through a two-stage optimization, Joint-GRPO trains the VLM to produce captions that are both accurate and friendly to visualize, while guiding the VDM to generate videos that are faithful to these captions and the input visual context.

To enable this learning, we construct VANS-Data-100K, a dedicated dataset with 100K video-question-answer triplets for supervised fine-tuning (SFT) on VNEP task. From this collection, we manually select 1K high-quality samples to support the subsequent RL post-training.

Experimental results on procedural and predictive VNEP benchmarks demonstrate that VANS performs favorably against state-of-the-art (SOTA) approaches in both event prediction accuracy and the quality of the generated videos. 

We make the following contributions in this work:
\begin{itemize}
\item We pioneer \textbf{VNEP}, advancing next-event reasoning from textual description to dynamic video demonstration.

\item We propose \textbf{VANS} and its core \textbf{Joint-GRPO} strategy, which aligns a VLM and a VDM through RL with a joint reward, yielding video answers that are both semantically faithful and visually coherent.

\item We construct \textbf{VANS-Data-100K}, a dataset of 100K (input video, question, output video) triplets for training and evaluating models on VNEP.

\end{itemize}

%% file: sec/2_related_works.tex
\section{Related Work}
\label{sec: related work}

\textbf{Next-Event Prediction.} The established NEP task requires predicting a future event given a video and a procedural or predictive question~\cite{cheng2025TEMPURA,lai2025cross,wang2025fostering,su2025eventformer}. Existing efforts predominantly address this as a textual NEP problem, focusing on generating descriptive answers. This line of work is supported by benchmarks like VLEP~\cite{lei2020more}, MVP~\cite{tan2023multiscale} and V1-33K~\cite{wang2025fostering}, and leverages techniques ranging from event understanding~\cite{chao2015hico} to multiscale temporal modeling and commonsense reasoning~\cite{chen2021joint}. The recent rise of VLMs has further propelled this field, with methods that fine-tune on large-scale data~\cite{cheng2025TEMPURA} or utilize RL to elicit NEP capabilities from pre-trained models~\cite{wang2025fostering}. However, a fundamental limitation persists: the answer modality remains exclusively textual. Our work breaks from this paradigm by introducing VNEP, advancing next-event reasoning from textual description to dynamic video demonstration.

\vspace{4pt} \noindent \textbf{Group Relative Policy Optimization.} Group Relative Policy Optimization (GRPO) was initially introduced by DeepSeek-Math~\cite{shao2024deepseekmath} to enhance the reasoning capabilities of language models and align their outputs with human preferences. Its effectiveness has led to its adoption in video understanding and reasoning~\cite{li2025videochat,wang2025video,feng2025video,chen2025grpo,ge2025arc,chen2025exploring}, where it improves model performance on complex, open-ended queries. Beyond understanding, GRPO has also been applied to image and video generation~\cite{jiang2025t2i,xiao2025mindomni,yan2025can,mi2025milr,xue2025dancegrpo}. In video generation, its primary role has been to enhance the alignment between the generated video and the text prompt~\cite{liu2025flow,xue2025dancegrpo}, as well as to improve consistency in reference-based generation tasks~\cite{shen2025identity}. While these works apply GRPO to optimize a single model, our Joint-GRPO coordinates two models (a VLM for reasoning and a VDM for visualization) simultaneously, ensuring they are jointly aligned for the VNEP task.

%% file: sec/3_dataset.tex
\section{VANS-Data-100K}
\label{sec: data}

Existing NEP datasets are unsuitable for direct use in VNEP due to suboptimal video quality and a lack of diverse instructional questions. To bridge this gap, we construct the VANS-Data-100K dataset, comprising 30K procedural and 70K predictive samples. Each sample contains an input video, a question, and a multi-modal answer (text and video), tailored for the VNEP task. As illustrated in Figure~\ref{fig: data pipeline}, our curation pipeline involves four stages.

\vspace{4pt} \noindent \textbf{Raw Data Collection.} We collect data from two distinct sources to cover both procedural and predictive scenarios. For procedural data, we source high-resolution videos from COIN~\cite{coin} and YouCook2~\cite{zhou2018towards} to ensure clear visual demonstrations of step-by-step tasks. For predictive data, we gather videos from general-scene datasets~\cite{wang2025fostering,caba2015activitynet} and short-films~\cite{cheng2025video}, which are rich in narrative and causal dynamics.

\vspace{4pt} \noindent \textbf{Shot Split.} Raw videos are segmented into coherent clips. Procedural videos are segmented using ground-truth timestamps, while predictive videos employ a shot-boundary detection model. We filter out segments shorter than 3 seconds to ensure action completeness.

\vspace{4pt} \noindent \textbf{Clip Selection.} We employ Gemini-2.5-Flash~\cite{team2023gemini} as an automated quality filter to identify the optimal 3-5 second clip. For procedural data, it selects the clip that best aligns with the given caption. For predictive data, it first generates a detailed caption for each segment, ensuring the selected clip is both high-quality and semantically representative.

\vspace{4pt} \noindent \textbf{QA Pair Generation.} Using Gemini-2.5-Flash, we generate QA pairs from video-caption sequences. The VLM simulates diverse questions—focusing on the logical next step for procedural tasks and hypothetical ``what-if'' scenarios for predictive ones. It also produces a chain-of-thought reasoning and ground-truth answer for each question, with a self-check to ensure logical soundness and prevent information leakage.

Please refer to Appendix~\ref{app: A} for more dataset details. 

%% file: sec/4_methods.tex
\section{VANS}
\label{sec: Method}

\begin{figure}[!t]
	\centering
	\includegraphics[width=0.4\textwidth]{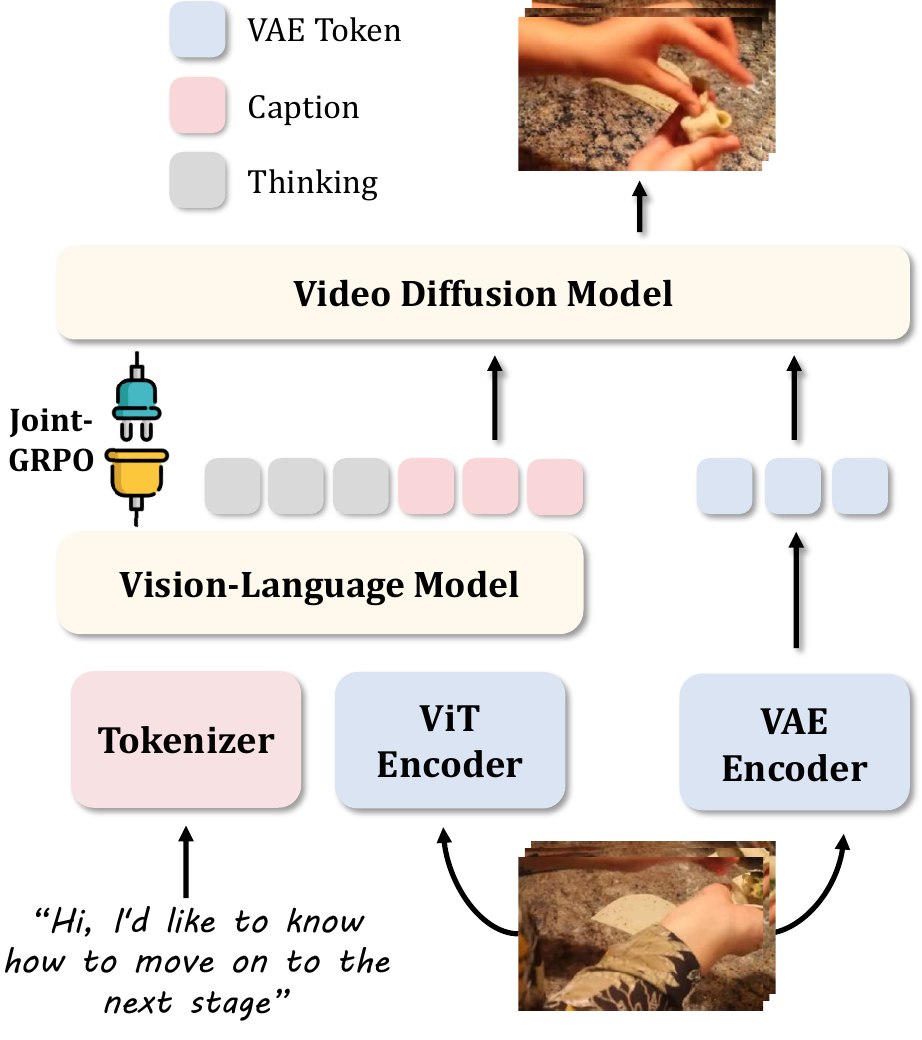}
\caption{Overall architecture of VANS.}
\label{fig: model architecture}
\end{figure}

Figure~\ref{fig: model architecture} introduces the overall architecture of VANS. The input question is tokenized and fed into the VLM alongside high-level ViT visual features from the input video. We task the VLM with performing instruction-grounded reasoning to generate a textual caption describing the predicted next event, which serves as the semantic guide for the VDM. To ensure visual consistency, the VDM is conditioned on both the generated caption and low-level visual cues, which are extracted by tokenizing $n$ sampled input frames using a VAE~\cite{wan2025wan}; these tokens are then concatenated into the VDM's conditioning latent space. This enables fine-grained visual correspondence while generating novel scenes.

This design faces a fundamental limitation: \textbf{the VLM and VDM are optimized in isolation}. The VLM is trained for textual accuracy but receives no feedback on whether its descriptions lead to visually plausible videos. Conversely, the VDM faces the challenge of coordinating two conditioning signals: the VLM's specific caption and the input's visual context. While SFT equips the VDM with basic capabilities, achieving consistent performance on both semantic accuracy and visual fidelity requires further refinement. This disconnect creates a \textit{semantic-to-visual gap} where both models operate without awareness of each other's constraints and capabilities. To resolve this, we introduce Joint-GRPO to orchestrate the two models into a cohesive unit for VNEP.

\begin{figure*}[!t]
	\centering
	\includegraphics[width=0.9\textwidth]{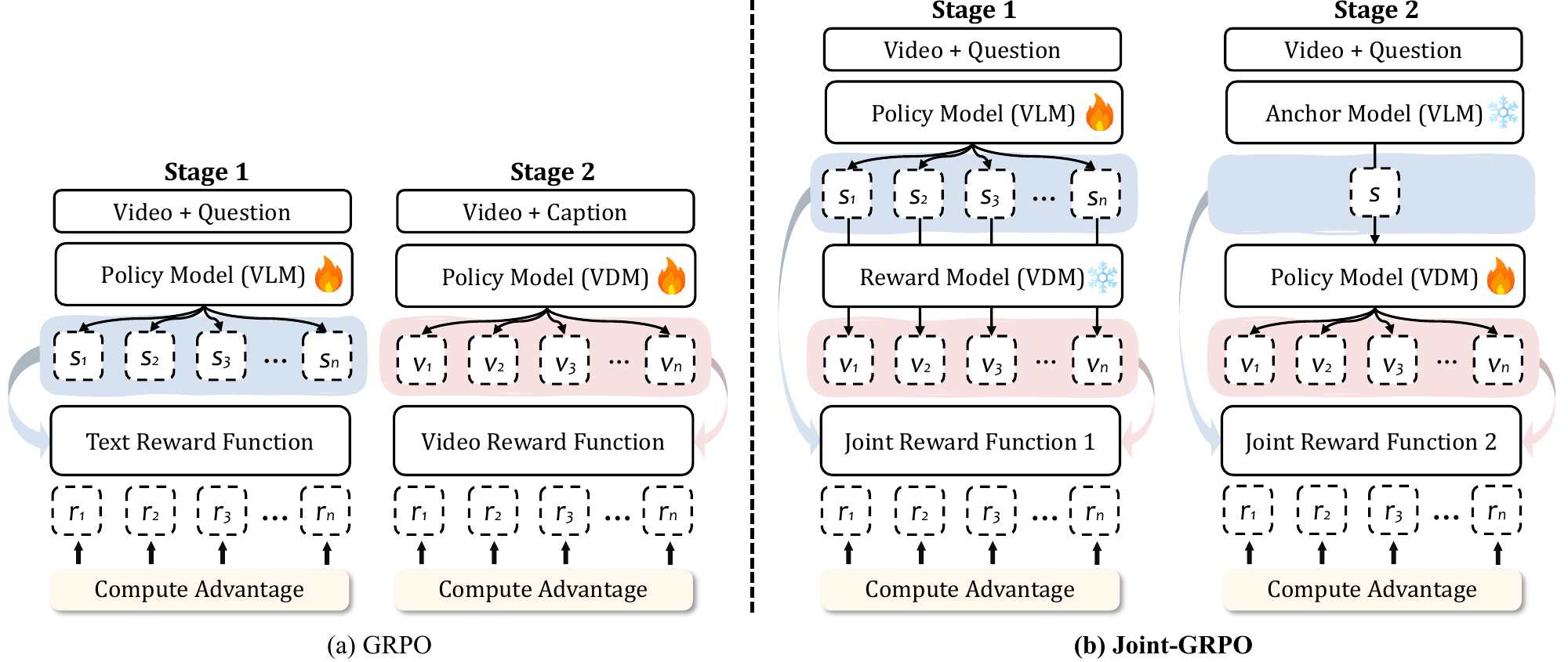}
\caption{Comparison of standard GRPO with Joint-GRPO. While standard GRPO optimizes a single model at a time, our Joint-GRPO coordinates their optimization under a joint reward function.}
\label{fig: joint grpo}
\end{figure*}

\subsection{Preliminary of GRPO}
GRPO is an RL algorithm designed to align model outputs with human preferences or complex objectives. The core concept involves using a \textit{reward function} to evaluate the quality of generated samples, and then adjusting the model's policy to increase the likelihood of high-reward generations. For each input context $c$, the policy model $\pi_\theta$ generates a group of $G$ trajectories $\{o_i\}_{i=1}^G$. Each trajectory receives a reward $r_i$ reflecting its quality. GRPO computes a normalized advantage $\tilde{A}_i$ that measures how much better or worse each trajectory is compared to the group average:
\begin{equation}
\tilde{A}_i = \frac{r_i - \bar{r}}{\sigma_r}, \quad
\bar{r} = \frac{1}{G}\sum_{j=1}^{G} r_j, \quad
\sigma_r = \sqrt{\frac{1}{G}\sum_{j=1}^{G} (r_j - \bar{r})^2}.
\end{equation}
The policy is then optimized using the GRPO objective:
\begin{multline}
J(\theta) = \mathbb{E} \left[ \frac{1}{G} \sum_{i=1}^{G} \left( \frac{1}{T_i} \sum_{t=0}^{T_i-1} \min\left( r_t^i(\theta) \tilde{A}_i, \right. \right. \right. \\
\left. \left. \left. \text{clip}(r_t^i(\theta), 1-\epsilon, 1+\epsilon) \tilde{A}_i \right) \right) - \beta D_{\text{KL}}(\pi_\theta \| \pi_{\text{ref}}) \right],
\label{eq:grpo}
\end{multline}
where \( r_t^i(\theta) = \frac{\pi_\theta(o_t^i | c)}{\pi_{\theta_{\text{old}}}(o_t^i | c)} \) is the probability ratio for the $i$-th trajectory. The clipping mechanism and KL divergence term ensure training stability by preventing drastic policy updates.

\subsection{Joint-GRPO}
\label{sec:joint-grpo}

Standard GRPO, while effective for single-model alignment, faces a fundamental limitation in multi-model scenarios like VNEP: it optimizes models in isolation. Applying it separately to the VLM and VDM fails to bridge the semantic-to-visual gap, as it does not encourage their outputs to be mutually reinforcing. Conversely, a one-stage joint training of both models is also problematic. This approach is prone to reward hacking and training instability, since when a generated video is of poor quality, it is ambiguous whether the VLM's caption or the VDM's generation is at fault, leading to conflicting gradient signals.

To address this attribution problem and enable effective co-steering, we propose Joint-GRPO. This approach coordinates the VLM and VDM using a joint reward function via a structured two-stage optimization process. Our key insight is that the two models must be co-steered such that the VLM's reasoning becomes visually grounded to guide the VDM effectively, while the VDM's generation remains faithful to the VLM's prediction and visual context.

\vspace{4pt} \noindent \textbf{Stage 1: Visualization-Friendly VLM Tuning.} We first align the VLM's reasoning with the VDM's generation results. We optimize the VLM policy \(\pi_{\text{VLM}}\) while keeping the VDM frozen. For an input video \(v_{\text{in}}\) and question \(Q\), we sample \(G\) textual captions \(\{s_i\}_{i=1}^G\) from \(\pi_{\text{VLM}}\). Each caption \(s_i\) is then used by the frozen VDM to generate a corresponding video \(v_{\text{out}}^i\). The joint reward \(r_1\) for the VLM is computed as:
\[
r_1(s_i, v_{\text{out}}^i) = \underbrace{\lambda_f r_f(s_i)}_{\text{format}} + \underbrace{\lambda_{t1} r_{t1}(s_i, s_{\text{gt}})}_{\text{text fidelity}} + \underbrace{\lambda_{v1} r_{v1}(v_{\text{out}}^i, v_{\text{gt}})}_{\text{video fidelity}},
\]
where \(\lambda_f, \lambda_{t1}, \lambda_{v1}\) are weighting coefficients for each reward, which are defined as follows:
\begin{itemize}
    \item \(r_f(s_i)\)  ensures the output follows the specified instruction format. A reward of 1 is given if the response adheres to the “reason-then-answer” template, and 0 otherwise.
    \item \(r_{t1}(s_i, s_{\text{gt}})\) measures semantic similarity between generated and ground-truth captions using ROUGE-L~\cite{lin2004rouge}.
    \item \(r_{v1}(v_{\text{out}}^i, v_{\text{gt}})\) evaluates visual coherence of generated videos with ground-truth using CLIP Similarity~\cite{radford2021learning}.
\end{itemize}

This composite reward is designed to steer the VLM beyond mere linguistic correctness. Relying solely on \(r_{t1}\) can lead to captions that are linguistically correct but visually unrealistic or unexecutable by the VDM. Conversely, using only \(r_{v1}\) provides a reward that is too distal and ambiguous to effectively guide the VLM's reasoning process. The joint reward guides the VLM to generate captions that are not merely semantically accurate, but also visually plausible and actionable for the VDM. This process effectively forces the VLM to internalize the VDM's capabilities and constraints.

\begin{table*}[!t]
  \caption{Quantitative comparison with baseline models on Video-Next-Event Prediction.}
  \label{tab: main results}
  \centering
\resizebox{0.83\textwidth}{!}{
\begin{tabular}{lccccc|ccc}
\toprule
\multicolumn{1}{c}{Model} & BELU@1$\uparrow$ & BELU@2$\uparrow$ & BELU@3$\uparrow$ & BELU@4$\uparrow$ & ROUGE-L$\uparrow$ & FVD$\downarrow$ & CLIP-V$\uparrow$ & CLIP-T$\uparrow$ \\
\midrule
\midrule
\vspace{-4mm} \\
\multicolumn{9}{>{\columncolor{LightBlue}}c}{Procedural Benchmarks} \\
\vspace{-2.8mm} \\
Video-GPT   & - & - & - & - & - & 105.32 & 0.7334 & 0.1997 \\
Omni-Video         & 0.0948  & 0.0253 & 0.0040  & 0.0008  & 0.1075 & 236.38  & 0.6293 & 0.2323 \\
Qwen-Wan           & 0.0981  & 0.0260  & 0.0046 & 0.0013  & 0.1530  & 148.75 & 0.6619 & 0.2448 \\
TEMPURA-Wan       & 0.1984  & 0.1063 & 0.0336 & 0.0167  & 0.1915 & 143.80   & 0.6738 & 0.2498 \\
Gemini-Wan         & 0.2432  & 0.1077 & 0.0448 & 0.0215  & 0.2802 & 120.34   & 0.6898 & 0.2547 \\
Qwen-FilmWeaver   & 0.0981  & 0.0260  & 0.0046 & 0.0013  & 0.1530  & 129.44 & 0.6831 & 0.2532 \\
TEMPURA-FilmWeaver & 0.1984  & 0.1063 & 0.0336 & 0.0167  & 0.1915 & 120.34   & 0.6923 & 0.2562 \\
Gemini-FilmWeaver   & 0.2432  & 0.1077 & 0.0448 & 0.0215  & 0.2802 & 110.54   & 0.7102 & 0.2773 \\
\textbf{VANS (SFT)}         & \uline{0.2524}  & \uline{0.1162} & \uline{0.0501} & \uline{0.0233}  & \uline{0.2812} & \uline{85.34}   & \uline{0.7655} & \uline{0.3202} \\
\textbf{VANS (Joint-GRPO)}  & \textbf{0.3257}  & \textbf{0.1834} & \textbf{0.1242} & \textbf{0.0987}  & \textbf{0.3631} & \textbf{78.32}   & \textbf{0.8021} & \textbf{0.3824} \\
\bottomrule
\vspace{-2.8mm} \\
\multicolumn{9}{>{\columncolor{color3}}c}{Predictive Benchmarks} \\
\vspace{-2.8mm} \\
Video-GPT   & - & - & - & - & - & 170.32 & 0.7031 &  0.2124 \\
Omni-Video         & 0.0885  & 0.0232 & 0.0035  & 0.0006  & 0.1012 & 252.47  & 0.6083 & 0.2218 \\
Qwen-Wan           & 0.0927  & 0.0241  & 0.0040 & 0.0010  & 0.1453  & 158.92 & 0.6427 & 0.2349 \\
TEMPURA-Wan        & 0.1639  & 0.0647 & 0.0132 & 0.0105  & 0.2142 & 152.86   & 0.6524 & 0.2398 \\
Gemini-Wan         & 0.1981  & 0.0760 & 0.0182 & 0.0112  & 0.2298 & 128.65   & 0.6673 & 0.2446 \\
Qwen-FilmWeaver    & 0.0927  & 0.0241  & 0.0040 & 0.0010  & 0.1453  & 137.84 & 0.6608 & 0.2431 \\
TEMPURA-FilmWeaver & 0.1839  & 0.0647 & 0.0132 & 0.0105  & 0.2142 & 128.32   & 0.6709 & 0.2463 \\
Gemini-FilmWeaver   & 0.1981  & 0.0760 & 0.0182 & 0.0112  & 0.2298 & 118.27   & 0.6874 & 0.2663 \\
\textbf{VANS (SFT)}  & \uline{0.2247}  & \uline{0.0873} & \uline{0.0206} & \uline{0.0136}  & \uline{0.2435} & \uline{94.12}   & \uline{0.7512} & \uline{0.3038} \\
\textbf{VANS (Joint-GRPO)}  & \textbf{0.2789}  & \textbf{0.1351} & \textbf{0.0853} & \textbf{0.0694}  & \textbf{0.3058} & \textbf{86.85}  & \textbf{0.7872} & \textbf{0.3759} \\
\bottomrule
\end{tabular}
}
\end{table*}

\vspace{4pt} \noindent \textbf{Stage 2: Context-Faithful VDM Adaptation.} 
Building upon the visually-grounded captions from Stage 1, this stage tackles the challenge of cross-modal alignment by adapting the VDM to render these captions faithfully while preserving visual consistency with the input visual context. We optimize the VDM policy \(\pi_{\text{VDM}}\) using the VLM as a frozen anchor model. As shown in Figure~\ref{fig: joint grpo}, the `now-improved' VLM from Stage 1 generates a candidate anchor caption (samples with low semantic similarity to the ground truth are discarded and regenerated to ensure quality). The resulting semantically-grounded caption \(s_{\text{anchor}}\) is then used to condition the VDM. We then sample \(G\) output videos \(\{v_{\text{out}}^i\}_{i=1}^G\) from \(\pi_{\text{VDM}}\). The VDM's core task is to generate a novel scene by dynamically attending to and preserving relevant visual elements (e.g., IDs, backgrounds) from the input video's VAE tokens, as guided by the semantic content of \(s_{\text{anchor}}\). The reward function \(r_2\) is defined as:
\[
r_2(v_{\text{out}}^i, s_{\text{anchor}}) = \underbrace{\lambda_{v2} r_{v2}(v_{\text{out}}^i, v_{\text{gt}})}_{\text{video fidelity}} + \underbrace{\lambda_{c2} r_{c2}(v_{\text{out}}^i, s_{\text{anchor}})}_{\text{semantic alignment}},
\]
where \(\lambda_{v2}, \lambda_{c2}\) are balancing coefficients, and:
\begin{itemize}
    \item \(r_{v2}(v_{\text{out}}^i, v_{\text{gt}})\) maintains visual quality and coherence with the input video, using the same metric as in Stage 1.
    \item \(r_{c2}(v_{\text{out}}^i, s_{\text{anchor}})\) measures semantic consistency between the output video and the anchor caption using CLIPScore.
\end{itemize}

This joint-reward design tackles the core challenge of cross-modal alignment. \(r_{v2}\) ensures the output remains visually plausible and continuous. \(r_{c2}\) compels the VDM to strictly adhere to the event described in \(s_{\text{anchor}}\), preventing it from ignoring the caption and merely reconstructing or slightly altering the input video.

Through this two-stage optimization, the VLM and VDM co-evolve into a synergistic unit. The distinct, complementary roles of each reward component, along with the training reward curves, are presented in Appendix~\ref{app: B}.

%% file: sec/5_exp.tex
\section{Experiments}
\label{sec: exp}
We conduct experiments to evaluate the effectiveness of our VANS and to compare it with cutting-edge solutions.

\subsection{Settings}

\vspace{4pt} \noindent \textbf{Benchmarks.} We construct our evaluation benchmark by sampling 400 procedural and 400 predictive samples from our dataset with source videos from established benchmarks~\cite{zhou2018towards,coin,caba2015activitynet,cheng2025video} to ensure reliable ground-truth text \& video answers. The evaluation set is strictly separated from the training data, with no video or question overlap.

\vspace{4pt} \noindent \textbf{Metrics.} Following~\cite{lai2025cross}, we employ BELU@1/2/3/4~\cite{papineni2002bleu} and ROUGE-L~\cite{lin2004automatic} for textual prediction quality. For videos, we use Fréchet Video Distance (FVD)~\cite{unterthiner2019fvd}, CLIP-Video Score (CLIP-V)$\uparrow$~\cite{radford2021learning}, and CLIP-Text Score (CLIP-T)~\cite{radford2021learning} to assess visual quality and semantic alignment.

\vspace{4pt} \noindent \textbf{Baselines.}
As no existing methods are designed for VNEP, we adapt top-performing models from related fields to establish baselines. These include: (1) Video extension model Video-GPT~\cite{zhuang2025videogptclipdiffusion}; (2) Cascaded pipelines formed by combining top-tire VLMs (Gemini-2.5-Flash~\cite{team2023gemini}, Qwen-2.5-VL-3B~\cite{bai2023qwen}, and its NEP-finetuned version TEMPURA~\cite{cheng2025TEMPURA}) with VDMs (Wan-2.1-1.3B~\cite{wan2025wan}, FilmWeaver~\cite{FilmWeaver}); and (3) Unified model Omni-Video~\cite{tan2025omni}.

\begin{figure*}[!t]
	\centering
	\includegraphics[width=0.84\textwidth]{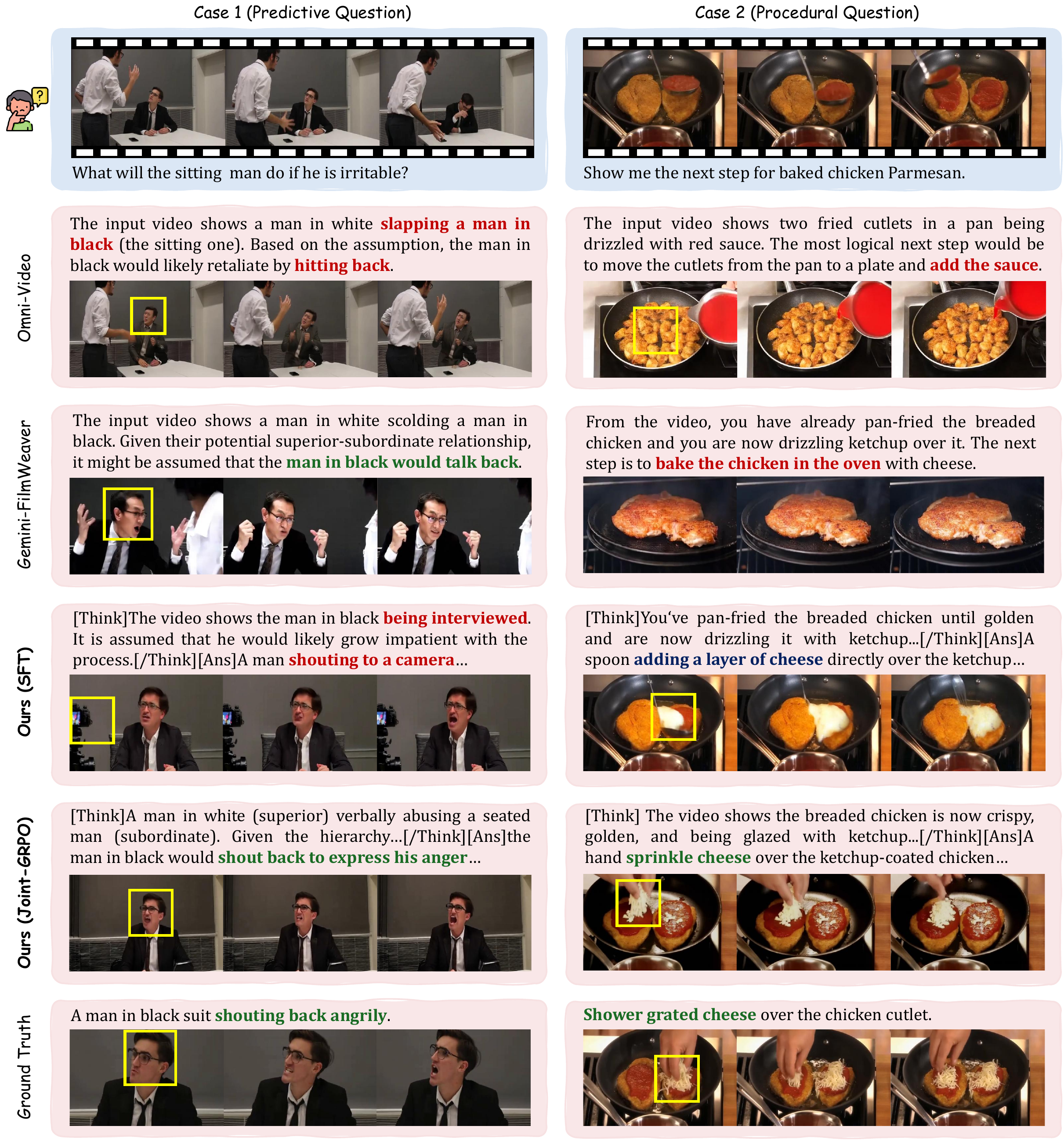}
\caption{Visual comparison on VNEP. Captions are color-coded: \textcolor{vis3}{green} (correct), \textcolor{vis1}{red} (incorrect), \textcolor{vis2}{blue} (semantically correct but visually unfriendly). Yellow boxes highlight key regions. Baselines often fail in event prediction or visual consistency. Our SFT model improves reasoning but retains errors like semantic hallucination (predicting non-existent inreview in Case 1) and action misalignment (``adding cheese'' yields pouring in Case 2). Joint-GRPO addresses both issues, enhancing model capability (correctly identifying document relationships and maintaining character appearance in Case 1) and fine-grained alignment (``sprinkle cheese'' matching the GT ``shower'' in Case 2).}
\label{fig: visualization}
\end{figure*}

\vspace{4pt} \noindent \textbf{Implementation Details.} 
We initialize VANS with Qwen2.5-VL-3B as the VLM and Wan-2.1-1.3B as the VDM. For Video-GPT, we provide the input video and utilize its native capability for video continuation. For VANS and other baseline methods, we supply the input video and the corresponding question to perform NVEP. Implementation details of our VANS and baselines are presented in Appendix~\ref{app: C}.

\begin{table*}[!t]
  \caption{Quantitative results of ablation study.}
  \label{tab: ablation results}
  \centering
\resizebox{0.81\textwidth}{!}{
\begin{tabular}{lccccc|ccc}
\toprule
\multicolumn{1}{c}{Model} & BELU@1$\uparrow$ & BELU@2$\uparrow$ & BELU@3$\uparrow$ & BELU@4$\uparrow$ & ROUGE-L$\uparrow$ & FVD$\downarrow$ & CLIP-V$\uparrow$ & CLIP-T$\uparrow$ \\
\midrule
\midrule
\vspace{-4mm} \\
\multicolumn{9}{>{\columncolor{LightBlue}}c}{Procedural Benchmarks} \\
\vspace{-2.8mm} \\
SFT                                   & 0.2524 & 0.1162 & 0.0501 & 0.0233 & 0.2812 & 85.34 & 0.7655 & 0.3202 \\
GRPO (VLM)                            & 0.2831 & 0.1498 & 0.0987 & 0.0698 & 0.3190 & 83.88 & 0.7798 & 0.3224 \\
GRPO (VDM)                            & 0.2524 & 0.1162 & 0.0501 & 0.0233 & 0.2812 & 84.76 & 0.7671 & 0.3013 \\
GRPO (VLM+VDM)                        & 0.2831 & 0.1498 & 0.0987 & 0.0698 & 0.2894 & 83.14 & 0.7703 & 0.3398 \\
Joint-GRPO Stage 1                    & \textbf{0.3257} & \textbf{0.1834} & \textbf{0.1242} & \textbf{0.0987} & \textbf{0.3631} & 80.23 & 0.7803 & 0.3521 \\
Joint-GRPO Stage 1 (w/o $r_{t1}$)  & 0.3176 & 0.1623 & 0.1123 & 0.0889 & 0.3498 & 83.31 & 0.7762 & 0.3454 \\
Joint-GRPO Stage 1 (w/o $r_{v1}$) & 0.3252 & 0.1828 & 0.1240 & 0.0978 & 0.3625 & 82.34 & 0.7668 & 0.3403 \\
Joint-GRPO Stage 1 + 2 (w/o $r_{c2}$)   & \textbf{0.3257} & \textbf{0.1834} & \textbf{0.1242}& \textbf{0.0987} & \textbf{0.3631} & 78.55 & 0.7921 & 0.3673 \\
Joint-GRPO Stage 1 + 2 (w/o $r_{v2}$) & \textbf{0.3257} & \textbf{0.1834} & \textbf{0.1242} & \textbf{0.0987} & \textbf{0.3631} & 79.76 & 0.7887 & 0.3806 \\
\textbf{Joint-GRPO Stage 1 + 2 (Ours)}   & \textbf{0.3257} & \textbf{0.1834} & \textbf{0.1242} & \textbf{0.0987} & \textbf{0.3631} & \textbf{78.32} & \textbf{0.8021} & \textbf{0.3824} \\
\textcolor{gray}{Joint-GRPO (all-in-one)} & \textcolor{gray}{0.3012} & \textcolor{gray}{0.1773} & \textcolor{gray}{0.1003} & \textcolor{gray}{0.0632} & \textcolor{gray}{0.3577} & \textcolor{gray}{81.01} & \textcolor{gray}{0.7800} & \textcolor{gray}{0.3423} \\
\bottomrule
\vspace{-2.8mm} \\
\multicolumn{9}{>{\columncolor{color3}}c}{Predictive Benchmarks} \\
SFT                                   & 0.2247 & 0.0873 & 0.0206 & 0.0136 & 0.2435 & 94.12 & 0.7512 & 0.3038 \\
GRPO (VLM)                            & 0.2521 & 0.1124 & 0.0412 & 0.0289 & 0.2758 & 92.54 & 0.7643 & 0.3218 \\
GRPO (VDM)                            & 0.2247 & 0.0873 & 0.0206 & 0.0136 & 0.2435 & 93.45 & 0.7525 & 0.3051 \\
GRPO (VLM+VDM)                        & 0.2521 & 0.1124 & 0.0412 & 0.0289 & 0.2552 & 91.83 & 0.7558 & 0.3342 \\
Joint-GRPO Stage 1                    & \textbf{0.2789} & \textbf{0.1351} & \textbf{0.0853} & \textbf{0.0694} & \textbf{0.3058} & 89.92 & 0.7654 & 0.3462 \\
Joint-GRPO Stage 1 (w/o $r_{t1}$)  & 0.2718 & 0.1203 & 0.0765 & 0.0627 & 0.2934 & 92.11 & 0.7613 & 0.3396 \\
Joint-GRPO Stage 1 (w/o $r_{v1}$) & 0.2785 & 0.1346 & 0.0851 & 0.0685 & 0.3052 & 91.15 & 0.7529 & 0.3345 \\
Joint-GRPO Stage 1 + 2 (w/o $r_{c2}$)   & \textbf{0.2789} & \textbf{0.1351} & \textbf{0.0853} & \textbf{0.0694} & \textbf{0.3058} & 88.36 & 0.7772 & 0.3612 \\
Joint-GRPO Stage 1 + 2 (w/o $r_{v2}$) & \textbf{0.2789} & \textbf{0.1351} & \textbf{0.0853} & \textbf{0.0694} & \textbf{0.3058} & 89.57 & 0.7738 & 0.3642 \\
\textbf{Joint-GRPO Stage 1 + 2 (Ours)}                           & \textbf{0.2789} & \textbf{0.1351} & \textbf{0.0853} & \textbf{0.0694} & \textbf{0.3058} & \textbf{86.85} & \textbf{0.7872} & \textbf{0.3759} \\
\textcolor{gray}{Joint-GRPO (all-in-one)} & \textcolor{gray}{0.2574} & \textcolor{gray}{0.1308} & \textcolor{gray}{0.0684} & \textcolor{gray}{0.0452} & \textcolor{gray}{0.3012} & \textcolor{gray}{90.82} & \textcolor{gray}{0.7651} & \textcolor{gray}{0.3365} \\
\bottomrule
\end{tabular}
}
\end{table*}

\begin{figure*}[!t]
\centering
\includegraphics[width=0.86\textwidth]{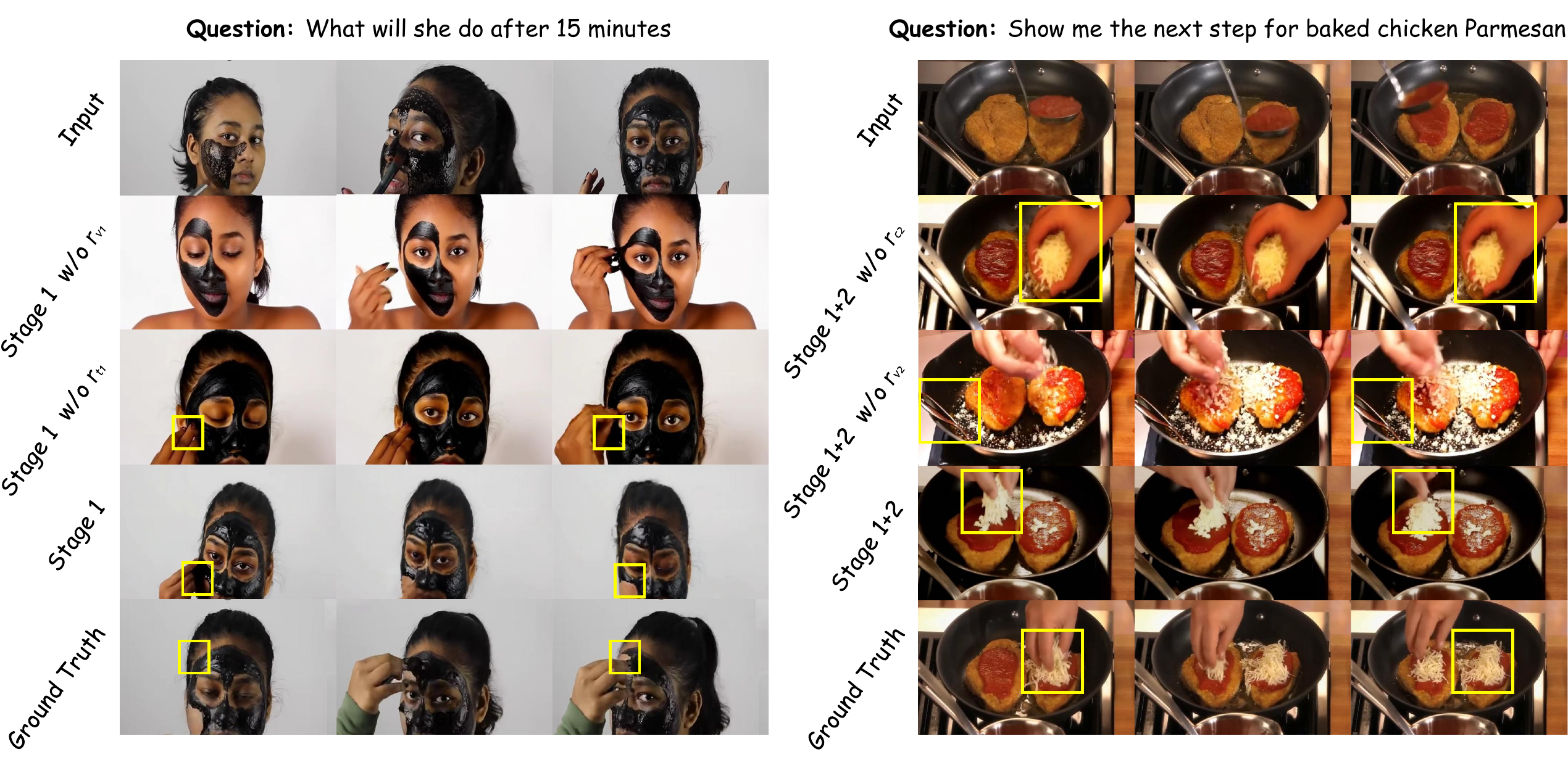}
\caption{Visualization Results of ablation studies. Key regions are highlighted with yellow boxes: the left example shows degradation in the ``mask removal" action completion without $r_{t1}$; the right example illustrates the loss of visual consistency without $r_{v2}$ and semantic alignment (leading to static frames) without $r_{c2}$.}
\label{fig: grpo teaser}
\end{figure*}

\subsection{Main Results}

\vspace{4pt} \noindent \textbf{Quantitative Comparisons.} Table~\ref{tab: main results} shows that VANS performs favorably against all baselines. On procedural benchmarks, VANS (Joint-GRPO) achieves a ROUGE-L of 0.3631 and CLIP-V of 0.8021, outperforming the strongest cascaded baseline (Gemini-FilmWeaver at 0.2802 and 0.7102) and unified model (Omni-Video at 0.1075 and 0.6293). More importantly, Joint-GRPO brings a significant gain over the SFT version (e.g., ROUGE-L from 0.2812 to 0.3631 and CLIP-V from 0.7655 to 0.8021), demonstrating the effectiveness of our Joint-GRPO strategy. The Video Extension model Video-GPT yields the lowest CLIP-T (0.1997), as it generates frames without event reasoning. Please refer to Appendix~\ref{app: D} for additional results.

\vspace{4pt} \noindent \textbf{Qualitative Comparisons.} As shown in Figure~\ref{fig: visualization}, baseline models frequently produce errors in either event prediction or visual consistency. For instance, Omni-Video misinterprets a quarrel as a fight and generates characters that deviate from the input. Our VANS after SFT demonstrates improved reasoning but reveals two key limitations: individual component errors, such as the VLM hallucinating non-existent text like ``inreview'' in Case 1, and a semantic-visual misalignment where the instruction ``adding cheese'' results in a pouring action instead of the ground-truth ``showering'' in Case 2. VANS with Joint-GRPO enhances the capability of each component and achieves semantic-visual alignment, as evidenced by the precise caption ``sprinkle cheese'' and its faithful visualization that matches the ``shower'' action.

\subsection{Ablation Study} 

We conduct ablation studies to validate the design of Joint-GRPO, with results in Table~\ref{tab: ablation results} and Figure~\ref{fig: grpo teaser}.

\vspace{4pt} \noindent \textbf{Joint vs. Isolated Optimization.}
Joint-GRPO outperforms variants where GRPO is applied solely to the VLM or VDM, or where their individually optimized versions are simply cascaded. This confirms the necessity of joint optimization for coherent caption-video generation, where the VLM and VDM are co-adapted to bridge the semantic-to-visual gap.

\vspace{4pt} \noindent \textbf{Effect of Staged Training.}
The two-stage design proves critical: using only Stage 1 often produces captions and videos that deviate semantically, while an all-in-one variant suffers from optimization instability due to reward ambiguity—it becomes unclear whether a poor reward stems from the VLM's caption or the VDM's video generation.

\vspace{4pt} \noindent \textbf{Reward Component Analysis.}
Further ablation tests validate the contribution of each reward component. In Stage 1, removing the text fidelity reward \(r_{t1}\) reduces caption accuracy (e.g., failing to predict ``removing the mask"), while removing the video fidelity reward \(r_{v1}\) harms visual consistency. In Stage 2, removing the semantic alignment reward \(r_{c2}\) causes reward hacking with static frames, and removing the video fidelity reward \(r_{v2}\) reduces output coherence. These findings validate our complete design with staged optimization and balanced reward components.

%% file: sec/6_Conclusion.tex
\section{Conclusion}
This work pioneers Video-Next-Event Prediction (VNEP), a novel task that advances next-event reasoning from textual description to dynamic video demonstration. To address its unique challenges, we introduce VANS, which synergizes a VLM and a VDM through Joint-GRPO—a two-stage RL strategy coordinating both models under a joint reward. We construct VANS-Data-100K dataset to provide the essential training and evaluation foundation for this task. Experiments on established benchmarks demonstrate that VANS achieves SOTA performance in both event prediction accuracy and video generation quality. 

\section*{Acknowledgement}
This work was supported by Kuaishou Technology.

%% file: sec/X_suppl.tex
\clearpage
\setcounter{page}{1}
\maketitlesupplementary

This Appendix is organized as follows:

\begin{itemize}
\item Section~\ref{app: A} provides the dataset details.
\item Section~\ref{app: B} offers an intuitive illustration of the Joint-GRPO reward design, along with training details.
\item Section~\ref{app: C} describes the implementation details of models.
\item Section~\ref{app: D} reports additional experimental results.
\end{itemize}

\section{Details of VANS-Data-100K}
\label{app: A}

Table~\ref{tab:dataset_details} demonstrates the composition and key statistics of our VANS-Data-100K dataset. It contains a total of 100K samples, with 30K dedicated to procedural tasks and 70K to predictive scenarios. These are sourced from a diverse set of publicly available video datasets and Internet to ensure broad coverage of real-world dynamics and instructional content.

In terms of video characteristics, the input videos have an average duration of 9.43 seconds, providing sufficient context for event reasoning. The corresponding target videos, which depict the predicted next event, average 3.76 seconds in length, ensuring concise and focused demonstrations.

\section{Details of Joint-GRPO}
\label{app: B}

\subsection{Reward Design}

\begin{figure*}[!t]
	\centering
	\includegraphics[width=0.87\textwidth]{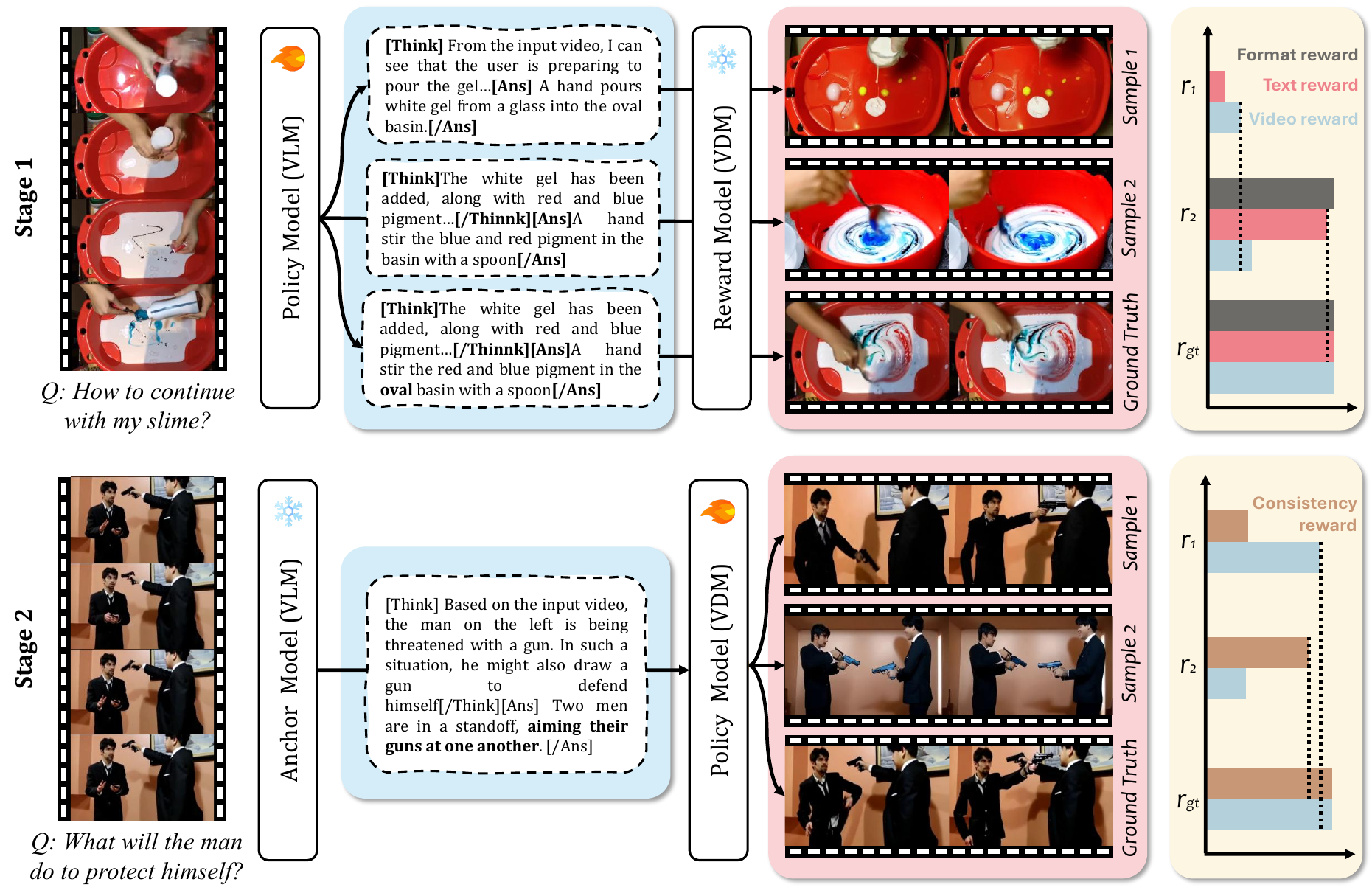}
\caption{Illustration of our Joint-GRPO reward design. Top: For a Stage-1 case, we simulate three reasoning samples during GRPO training. The text-only reward fails to penalize Sample 2's visual inconsistency, while the video-only reward fails to penalize Sample 1's semantic error. Bottom: For a Stage-2 case, the video-only reward fails to penalize Sample 1's semantic inaccuracy, while the consistency-only reward fails to penalize Sample 2's visual inconsistency.}
\label{fig: grpo reward design}
\end{figure*}

Figure~\ref{fig: grpo reward design} provides an intuitive illustration of how the individual reward components work in concert within our two-stage training process of Joint-GRPO.

In Stage 1 (VLM Tuning), we examine the role of the text fidelity reward (\(r_{t1}\)) and the video fidelity reward (\(r_{v1}\)). For the provided example, if only \(r_{t1}\) is used, Sample 2 receives a high score comparable to the Ground Truth (GT), as both captions correctly describe the action. However, Sample 2 exhibits poor visual consistency. Conversely, if only \(r_{v1}\) is used, both Sample 1 and Sample 2 receive similarly low scores, failing to reflect that Sample 1 is semantically worse due to an incorrect action prediction. Only when both \(r_{t1}\) and \(r_{v1}\) are combined does the composite reward correctly rank the samples, successfully identifying the GT as the best.

In Stage 2 (VDM Adaptation), we analyze the video fidelity reward (\(r_{v2}\)) and the semantic consistency reward (\(r_{c2}\)). Relying solely on \(r_{v2}\) results in Sample 1 receiving a score similar to the GT, even though Sample 1 depicts an incorrect semantic action (it should show two people pointing guns at each other). Using only \(r_{c2}\) causes Sample 2 to be scored similarly to the GT, despite its poor visual consistency. The joint reward effectively combines these signals to prioritize samples that are correct in both semantics and visual quality.

The final combined reward in each stage is a sum of the normalized individual rewards. All weighting coefficients (\(\lambda\)) are set to 1, assigning equal importance to each objective. This design ensures a balanced optimization towards captions that are both semantically accurate and visually plausible (Stage 1), and videos that are both high-quality and semantically faithful (Stage 2).

\begin{table}[!t]
\centering
\caption{Statistics of VANS-Data-100K dataset.}
\label{tab:dataset_details}
\small
\begin{tabular}{lr}
\toprule
\multicolumn{1}{c}{\textbf{Component}} & \textbf{Size/Duration} \\
\midrule
\textbf{Data Composition} & \\
\ \ Procedural (Total: 30K) & \\
\ \ \ \ YouCook2~\cite{zhou2018towards} & 9K \\
\ \ \ \ COIN~\cite{coin} & 21K \\
\ \ Predictive (Total: 70K) & \\
\ \ \ \ Video-Holmes~\cite{cheng2025video} & 10K \\
\ \ \ \ ActivityNet~\cite{caba2015activitynet} & 20K \\
\ \ \ \ V1-33K~\cite{wang2025fostering} & 10K \\
\ \ \ \ YouTube Videos & 30K \\
\midrule
\textbf{Video Duration (Avg.)} & \\
\ \ Input Video & 9.43s \\
\ \ Target Video & 3.76s \\
\bottomrule
\end{tabular}
\end{table}

\begin{figure*}[!t]
	\centering
	\includegraphics[width=0.90\textwidth]{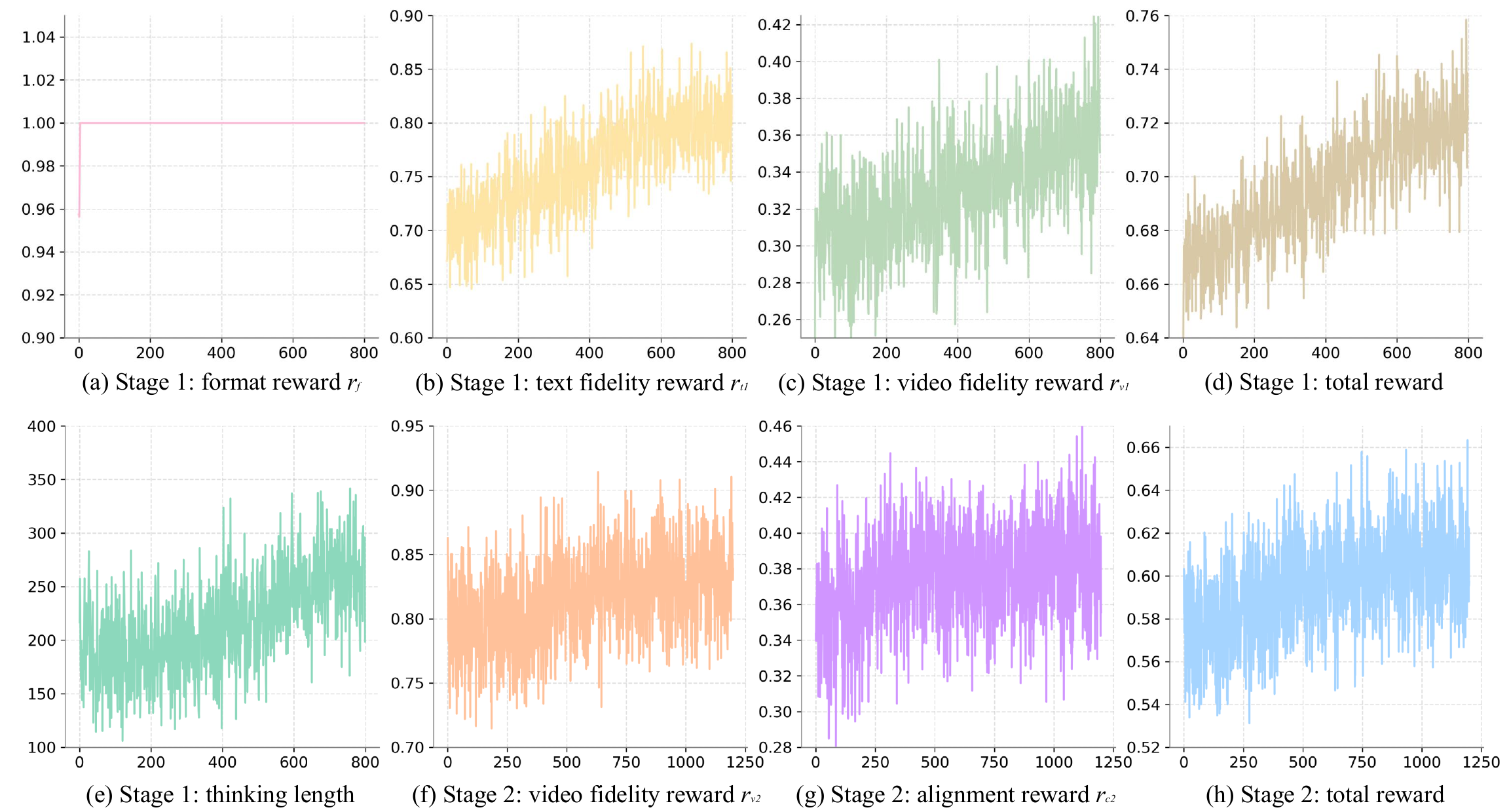}
	\caption{Training curves of Joint-GRPO: (a) format reward ($r_f$) in Stage 1; (b) text fidelity reward ($r_{t1}$) in Stage 1; (c) video fidelity reward ($r_{v1}$) in Stage 1; (d) total reward in Stage 1; (e) thinking length evolution in Stage 1; (f) video fidelity reward ($r_{v2}$) in Stage 2; (g) semantic alignment reward ($r_{c2}$) in Stage 2; (h) total reward in Stage 2.}
	\label{fig:training_curves}
\end{figure*}

\subsection{Training Process}

Figure~\ref{fig:training_curves} illustrates the training dynamics of Joint-GRPO. In Stage 1 (VLM Tuning), the format reward ($r_f$) in Figure~\ref{fig:training_curves}(a) quickly saturates, indicating rapid adoption of the instruction template. Both text fidelity ($r_{t1}$, Figure~\ref{fig:training_curves}b) and video fidelity ($r_{v1}$, Figure~\ref{fig:training_curves}c) rewards show progressive improvement, reflecting the VLM's learning to generate captions that are both semantically accurate and visually plausible. The combined reward (Figure~\ref{fig:training_curves}d) stabilizes after approximately 600 steps, demonstrating effective optimization. Concurrently, the increasing thinking length (Figure~\ref{fig:training_curves}e) indicates more detailed reasoning chains.

In Stage 2 (VDM Adaptation), both video fidelity ($r_{v2}$, Figure~\ref{fig:training_curves}f) and semantic alignment ($r_{c2}$, Figure~\ref{fig:training_curves}g) rewards improve consistently, with convergence occurring after about 1000 steps. This demonstrates the VDM's successful adaptation to generate videos that preserve visual consistency while faithfully rendering the semantically-grounded captions from Stage 1. The total reward (Figure~\ref{fig:training_curves}h) also reaches a stable level, confirming effective cross-modal alignment.

Collectively, these training curves validate the effectiveness of our Joint-GRPO design, demonstrating coordinated improvement across both stages.

\section{Implementation Details}
\label{app: C}

\subsection{Training of VANS}

We initialize VANS with Qwen2.5-VL-3B as the VLM and Wan-2.1-1.3B as the VDM. The VDM is configured to use \(n=6\) reference frames. 

In the SFT stage, the VLM is trained for 10K steps using LoRA~\cite{hu2022lora} (rank=8, alpha=32) with a learning rate of \(5\times10^{-5}\), while the VDM is fully fine-tuned for 20K steps across all DiT blocks with the same learning rate of \(5\times10^{-5}\).

For Joint-GRPO post-training, Stage 1 is optimized for 800 steps with a learning rate of \(5\times10^{-5}\). In Stage 2, to ensure the quality of anchor captions, we filter out those with ROUGE-L scores below 0.6 before proceeding with VDM adaptation. Stage 2 is then trained for 1K steps with the same learning rate of \(5\times10^{-5}\). We equip the VLM with LoRA (rank=8, alpha=32). For the VDM, we adopt the method from~\cite{liu2025flow} to convert a deterministic Ordinary Differential Equation (ODE) into an equivalent Stochastic Differential Equation (SDE) to enable GRPO training. We set the KL coefficient \(\beta = 0.004\), clip range to \(1\times10^{-3}\), and sample group size to 8 per prompt.

\subsection{Evaluation}

\noindent \textbf{Evaluation Protocol.} All methods are evaluated under a unified protocol. For Video-GPT, we provide only the input video and utilize its native video continuation capability. For VANS and other baseline methods, we provide the input video along with the corresponding question and the following system prompt:

\begin{tcolorbox}[colback=white,colframe=black,boxrule=1pt,arc=0pt]
You will be given a video. Your task is to predict the next event based on the input video and the user's instructions. Please begin by providing your detailed reasoning between the [Think][/Think] tags, followed by your detailed description of the next event within the [Ans][/Ans] tags.
\end{tcolorbox}

\noindent \textbf{Input Adaptation.} To accommodate different model architectures, we adapt the video input accordingly: for models that can directly process video input (e.g., Gemini), we provide the original video; for other models (e.g., Qwen), we set the input video fps = 1 for their video processors.

\noindent \textbf{Output Specification.} All methods are required to generate a video answer with a resolution of \(352 \times 640\) and a length of 33 frames to ensure consistent and fair comparison.

\noindent \textbf{Metric Computation.} The CLIP-Score for video consistency (CLIP-V) and semantic consistency (CLIP-T) is computed using a ViT-B/32 model. Specifically, each frame of the generated video is compared with the corresponding frame in the ground-truth video, and the scores are averaged across all frames.

\section{Additional Results}
\label{app: D}

\begin{figure*}[!t]
	\centering
	\includegraphics[width=\textwidth]{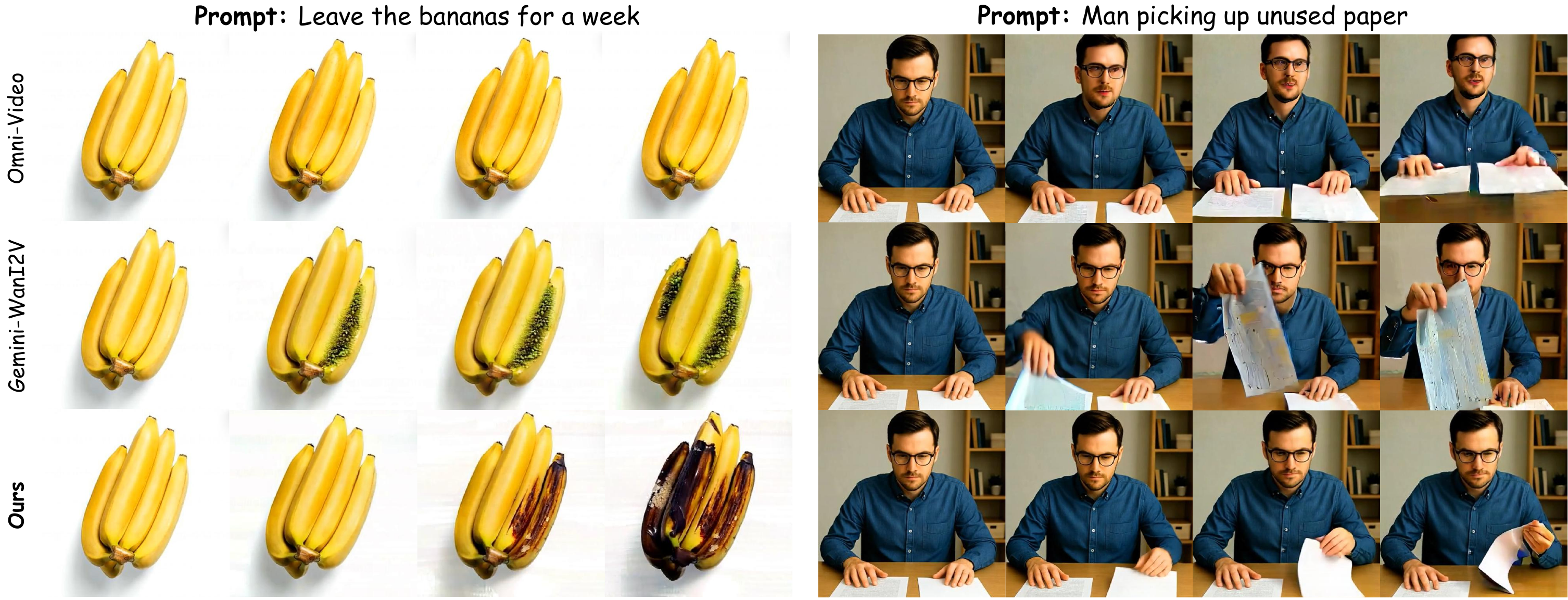}
\caption{Visual comparison results on UI2V-Bench.}
\label{fig: gen_ri2v}
\end{figure*}
\begin{figure}[!t]
	\centering
	\includegraphics[width=0.5\textwidth]{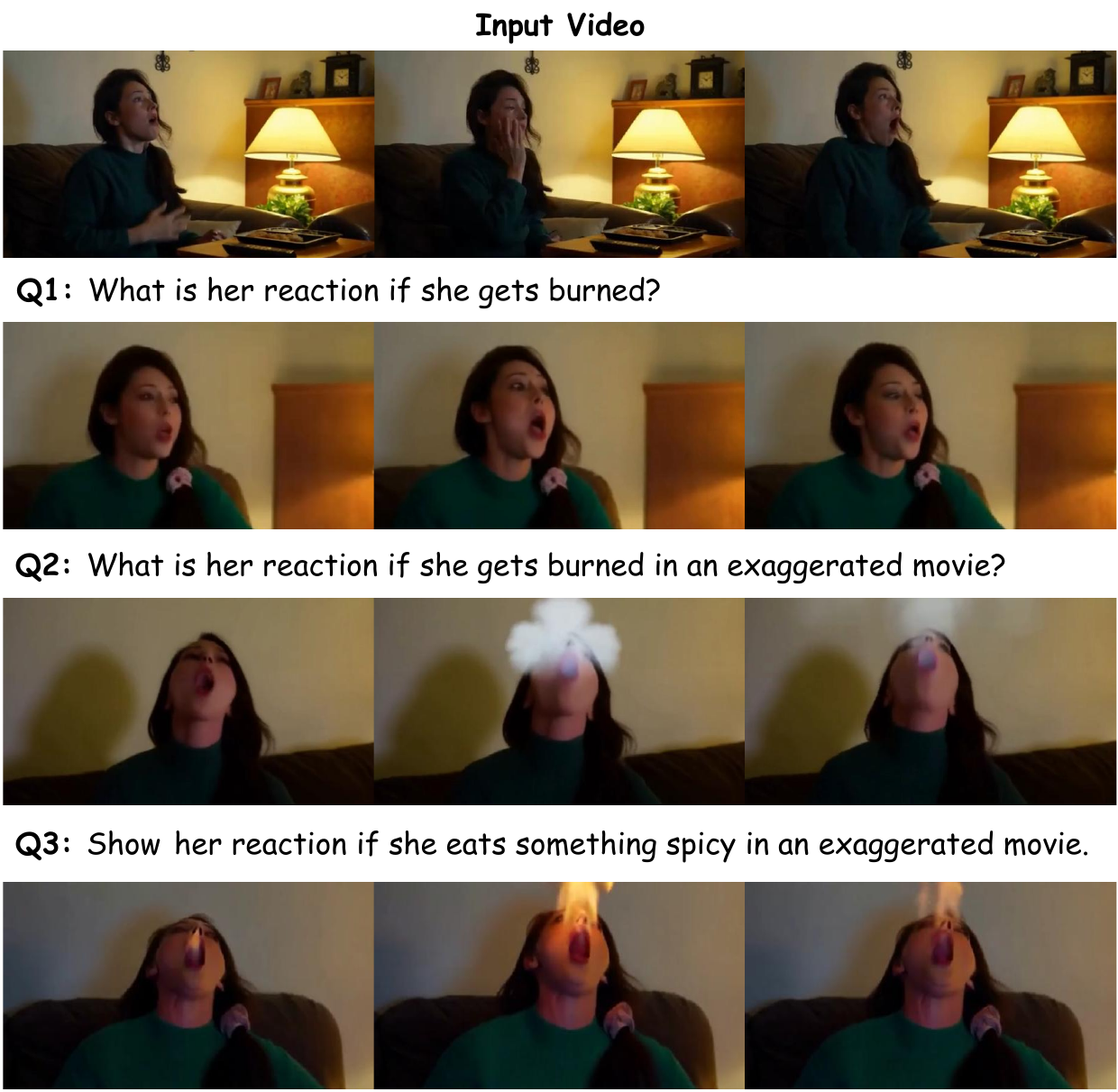}
\caption{Multi-Future Prediction Results.}
\label{fig: gen_Multi}
\end{figure}

\subsection{Inference Time}
The inference time of VANS is comparable to other cascaded pipelines, requiring approximately 4 seconds for caption generation and 35 seconds for video generation using the official VAN library. In contrast, unified models exhibit longer inference times: Omni-Video requires approximately 50 seconds, while VideoGPT needs about 60 seconds for complete generation.


\subsection{Comparison with Fine-tuned Baseline}

To analyze the source of performance improvements in VANS, we compare it with fine-tuned baselines. As shown in Table~\ref{tab:finetuned_comparison}, the results indicate three main observations: data quality provides a foundation, architectural modification contributes to noticeable gains, and Joint-GRPO provides the decisive enhancement that pushes performance to the state-of-the-art level.

\begin{table}[!t]
\caption{Results on procedural VNEP. The comparison with fine-tuned baselines (*) shows that our architectural design, rather than data advantage, is the primary source of improvement.}
\label{tab:finetuned_comparison}
\centering
\small
\resizebox{0.5\textwidth}{!}{
\begin{tabular}{lccccc}
\toprule
Model & BELU@4$\uparrow$ & ROUGE-L$\uparrow$ & FVD$\downarrow$ & CLIP-V$\uparrow$ & CLIP-T$\uparrow$ \\
\midrule
Qwen-Wan & 0.0013  & 0.1530  & 148.75 & 0.6619 & 0.2448  \\
Qwen*-Wan  & \uline{0.0233} & \uline{0.2812} & 140.32 & 0.6790 & 0.2466 \\
Qwen*-Wan*  & \uline{0.0233}  & \uline{0.2812}  & 140.07 & 0.6795 & 0.2470 \\
Gemini-Wan & 0.0215  & 0.2802 & 120.34   & 0.6898 & 0.2547 \\
\hline
\textbf{VANS (SFT)}  & \uline{0.0233}  & \uline{0.2812} & \uline{85.34}   & \uline{0.7655} & \uline{0.3202} \\
\textbf{VANS (Joint-GRPO)} & \textbf{0.0987}  & \textbf{0.3631} & \textbf{78.32}   & \textbf{0.8021} & \textbf{0.3824}  \\
\bottomrule
\end{tabular}
}
\end{table}

\vspace{4pt} \noindent \textbf{Data Quality as the Foundation.} When fine-tuned on our VANS-Data-100K for 10K steps (denoted as Qwen*), the model achieves reasoning capability competitive with Gemini-2.5-Flash in a zero-shot setting (ROUGE-L: 0.2812 vs. 0.2802). This confirms that our high-quality dataset enables smaller models to learn sophisticated reasoning.

\vspace{4pt} \noindent \textbf{Architectural Modification Contributes to Gains.} Fine-tuning both components of the Qwen-Wan pipeline (denoted as Qwen*-Wan*) yields limited video metric improvements over the base fine-tuned VLM (denoted as Qwen*-Wan). In contrast, VANS (SFT) with the same text input achieves better video results: FVD decreases from 140.07 to 85.34 and CLIP-V increases from 0.6795 to 0.7655, suggesting the proposed VAE reference feature aids visual consistency.

\vspace{4pt} \noindent \textbf{Joint-GRPO Delivers the Decisive Enhancement.} The most striking improvement comes from Joint-GRPO, which elevates VANS to unprecedented performance levels across all metrics. Compared to VANS (SFT), Joint-GRPO boosts ROUGE-L from 0.2812 to 0.3631 (29.1\% relative improvement) and CLIP-T from 0.3202 to 0.3824 (19.4\% relative improvement), while further reducing FVD to 78.32. These results unequivocally demonstrate that Joint-GRPO is the most critical component for achieving state-of-the-art performance, effectively aligning both textual and visual outputs with human preferences.

\subsection{Generalization}

\vspace{4pt} \noindent \textbf{Multi-Future Prediction.}
The established NEP task typically assumes a single, causal progression from the input context. In contrast, our VANS demonstrates a key generalization capability: multi-future prediction. This allows the model to generate semantically distinct and contextually appropriate video answers based on different hypothetical questions applied to the same input video, moving beyond deterministic continuation.

As shown in Figure~\ref{fig: gen_Multi}, when presented with a scene of a woman reacting to a hot object, VANS can generate fundamentally different yet plausible outcomes conditioned on the scenario: in a realistic everyday context, it predicts a natural reaction of ``coughing"; whereas in a stylized cinematic context, it visualizes a dramatic effect of ``smoke exhaling from the mouth". This flexibility stems from our model's ability to ground its predictions in both the visual evidence and the diverse textual hypotheses provided, effectively exploring multiple potential futures from a single starting point.

\vspace{4pt} \noindent \textbf{Reasoning Image-to-Video Generation.}
VANS generalizes effectively to the reasoning image-to-video (I2V) task by treating a single image as a static video clip. This generalization capability is attributed to the model's training on mixed datasets including Koala-36M~\cite{wang2024koala36mlargescalevideodataset} for I2V tasks. Figure~\ref{fig: gen_ri2v} demonstrates examples from UI2V-Bench~\cite{zhang2025ui2v}, when given an image of a banana and the instruction "leave the banana for a week," our model accurately predicts the temporal evolution, generating a video where the banana skin darkens. In contrast, other strong baselines struggle to capture this causal-physical transformation correctly. This demonstrates the robustness of our approach in understanding static visual contexts and reasoning about their potential dynamic futures.

\subsection{Human Evaluation}

To complement automatic metrics, we conduct a human evaluation to assess the subjective quality of generated video answers. We recruit 30 evaluators (mean age = 25 years; all hold at least a bachelor’s degree, including 20 postgraduate/PhD students and 10 full-time professionals). Each evaluator is presented with 20 randomly selected examples (10 procedural and 10 predictive) and rates the results on three dimensions: semantic correctness, visual consistency, and overall satisfaction.

The results in Table~\ref{tab:human_eval} reveal that VANS (SFT) achieves semantic correctness comparable to the strong baseline Gemini-FilmWeaver, while demonstrating superior visual consistency. Furthermore, VANS with Joint-GRPO receives the highest ratings across all three criteria, indicating that our full approach yields video answers that are not only semantically and visually accurate but also subjectively more satisfactory to human observers.

\subsection{Video Results}
All video results corresponding to the figures in this paper, along with additional examples, are provided in \url{supplementary.html}.


\begin{table}[!t]
\centering
\caption{Human evaluation results (scale: 1-5). Our VANS with Joint-GRPO achieves the highest scores across all criteria.}
\label{tab:human_eval}
\small
\resizebox{0.47\textwidth}{!}{
\begin{tabular}{lccc}
\toprule
\multicolumn{1}{c}{\textbf{Model}} & \textbf{Semantic Correctness} & \textbf{Visual Consistency} & \textbf{Overall} \\
\midrule
Video-GPT & 1.5 & 3.6 & 1.5 \\
Omni-Video & 2.1 & 3.2 & 2.2 \\
Gemini-FilmWeaver & 3.9 & 3.1 & 3.5 \\
VANS (SFT) & 3.8 & 3.9 & 3.7 \\
VANS (Joint-GRPO) & \textbf{4.7} & \textbf{4.6} & \textbf{4.8} \\
\bottomrule
\end{tabular}}
\end{table}